# Multi-model fusion for Aerial Vision and Dialog Navigation based on human attention aids

Xinyi Wang, Xuan Cui, Danxu Li, Fang Liu, Licheng Jiao
Xidian University, China

## Abstract

*Drones have been widely used in many areas of our daily lives. It relieves people of the burden of holding a controller all the time and makes drone control easier to use for people with disabilities or occupied hands. However, the control of aerial robots is more complicated compared to normal robots due to factors such as uncontrollable height. Therefore, it is crucial to develop an intelligent UAV that has the ability to talk to humans and follow natural language commands. In this report, we present an aerial navigation task for the 2023 ICCV Conversation History. Based on the AVDN dataset containing more than 3k recorded navigation trajectories and asynchronous human-robot conversations, we propose an effective method of fusion training of Human Attention Aided Transformer model (HAA-Transformer) and Human Attention Aided LSTM (HAA-LSTM) model, which achieves the prediction of the navigation routing points and human attention. The method not only achieves high SR and SPL metrics, but also shows a 7% improvement in GP metrics compared to the baseline model.*

## 1. Introduction

Drones are increasingly used in our daily lives, from personal entertainment to professional applications such as agriculture. Drones have the advantage of being more maneuverable than regular robots, but controlling aerial robots is also more complex because of the additional altitude control involved. Therefore, the control of UAV usually requires a human to use the controller, so I hope to find a way to control the movement of UAV without using hands according to the control experience of UAV users. That is, the UAV can obtain navigation instructions through dialogue with humans, so that users with disabled hands or users with occupied hands (such as painting, taking photos, writing, etc.) can also control the movement of the UAV. The Aerial Vision-and-Language Navigation Challenge[1] collected an AVDN dataset containing 3064 aerial navigation tracks and built a simulator with continuous properties. On this basis, the AVDN task is proposed, which presents a new challenge for the sequential action prediction under large continuous data, aiming at finding a model that can solve the problem better. Previous researches generally refer to Vision-and-Language Navigation as a multi-modal task. Specifically, the goal of VLN is to enable an agent to explore the unseen real environment according to natural language instructions and visual scenes, so as to realize specific tasks such as navigation and finding specific items. Anderson et al. based on the Matterport3D simulator (Chang et al., 2017). [2][3]The concrete agent in R2R moves through the house in the simulator, traverses the edges on the navigation chart, and jumps to adjacent nodes that contain the panoramic view. Krantz et al. [4]constructed a three-dimensional continuous environment to simulate continuous state changes in the real world by reconstructing scenes based on topological connections, in which agents use continuous actions during navigation. Other VLN research focuses on language instructions. Jain et al. [5] proposed that room-for-room connects paths in R2R to longer tracks. Yan et al. [6] collected xml-R2R and extended R2R with Chinese instructions. Anderson et al. [2] collected more data samples, in which the instructions were aligned with the virtual pose time of the instructions, and proposed RxR, containing the instructions in English, Hindi, and Telegu. He et al. [7] further extended the English split of RxR to Landmark RxR. Misra et al. [8] proposed LANI, a 3D synthetic navigation environment in which agents navigate between landmarks following natural language instructions. Because it contains more objects, outdoor environments are often more complex than indoor environments. The Touchdown dataset proposed by Chen et al. [9], the StreetLearn dataset proposed by Mirowski et al. [10], the StreeNav dataset proposed by Hermann et al. [11], and the Talk2Nav dataset proposed by Vasudevan et al. [12], are all based on Google Street View. However, this kind of data set is used in the synthetic environment, ignoring the control of the height of the UAV, and there is a certain gap with the real scene. This kind of navigation is overly simplified, and the AVDN dataset draws on the advantages of previous work to build continuous environments and conversational instructions that more closely resemble real-world scenarios.

For Aerial Navigation, there are also some works in this field that are useful for our research. Fan et al. [13] are committed to solving aerial visual navigation problems by using pre-collected real-world UAV data. Due to the difficulties of collecting real-world drone data, such as the risk of crash, Chen et al. [14] will apply simulation to air navigation.

Previous studies have provided a wealth of relevant knowledge, but previous work lacked the modality of language, so it could only perform simple navigation tasks. AVDN proposes navigation guided by natural dialogue, so that some complex and diverse navigation tasks can be performed.

## 2. Methods

We first use HAA-Transformer as a baseline model. The models take multimodal information as input, with three modes: drone orientation, drone visual observation images, and history dialog boxes. The output is to generate multimodal predictions, including human attention prediction and navigation prediction. To prevent the model from overfitting, we adjusted the number of iteration rounds of HAA-Transformer during training to find a balance between overfitting and underfitting. Since the task is also similar to time series prediction, we also tried HAA-LSTM and fused the model with the results of HAA-Transformer to select the optimal solution.

### 2.1. HAA-Transformer

HAA-Transformer takes multimodal information as input and generates multimodal predictions, including human attention predictions and navigation predictions. The multimodal coding input has three modes, the direction of the drone, the visual observation image of the drone, and the historical dialogue. At the beginning of the prediction series, HAA-Transformer uses BERT encoders to get language embeddings of the input conversation history, where special language tags such as [INS] and [QUE] are added before each instruction and question in the conversation. Then, at each time step, all previous drone directions and drone visual observations were fed into the model. All embeddings from language, image, and direction are wired and fed into a multimodal Transformer, producing the output multimodal embeddings.

## 3. Experiments

### 3.1. Experimental Setup

**Datasets.** We are using the AVDN dataset, the dataset contains 3064 air navigation tracks, each containing multiple rounds of conversations. On average, each trajectory contains only two rounds of dialogue, and the number of rounds of dialogue is the same as the maximum time step M. 3064 complete tracks can be divided into 6269 instructions and corresponding tracks. As for instructions, there are two types, one is a detailed description of the destination instruction, and the other is a rough initial instruction and further description of the subsequent dialogue. There were also two ways to describe direction, one self-centered, such as "turn right" (82 percent). One is non-self-centered, such as "turn south," which accounts for 30 percent. There are also some instructions that contain both types of instructions.

**Evaluation Metric.** The evaluation outputs include three metrics and SPL is the primary evaluation metric:

Success weighted by inverse Path Length (**SPL**) [15]: measuring the Success Rate weighted by the total length of the navigation trajectory.

Success Rate (**SR**): the number of the predicted trajectory being regarded as successful, i.e., the final view area of the predicted trajectory satisfies the IoU requirement, over the number of total trajectories predicted.

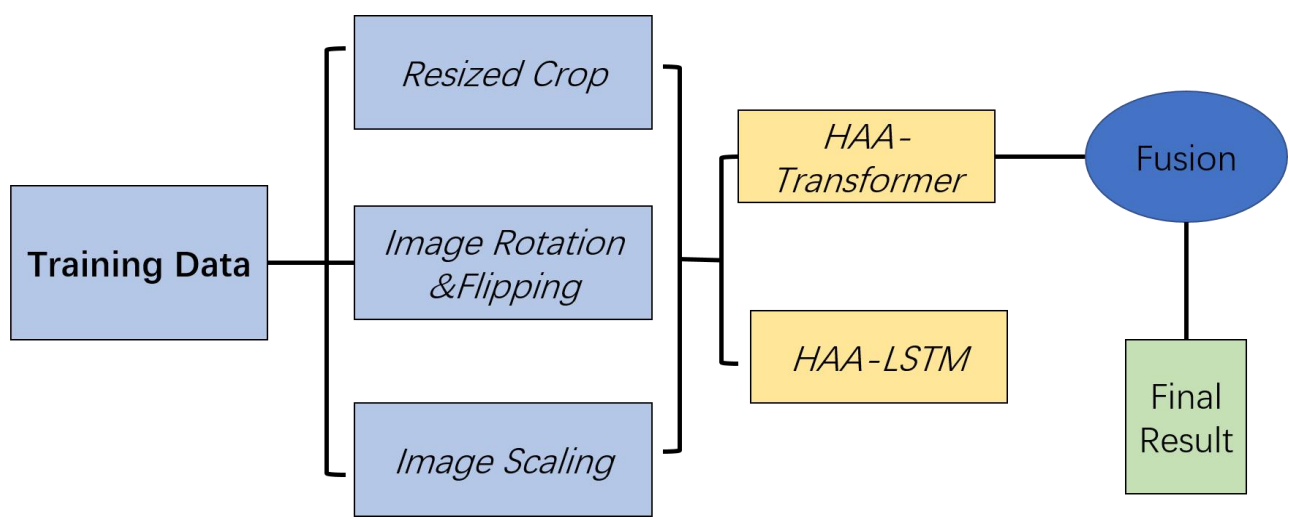

Figure 1: Detailed process of model training and fusion

Goal Progress (**GP**) [16]: evaluating the distance of the progress made towards the destination area. It is computed as the Euclidean distance of the trajectory, deducted by the remaining distance from the center of the predicted final view area to the center of goal area G.

**Implementation Details.** As shown in Fig. 1, it demonstrates our whole training process. Firstly, some simple data enhancements are performed on the ANDH dataset, including image blurring, random noise, random flipping, and so on. After 140k and 200k iterations of HAA-Transformer training, the results of different stages of HAA-Transformer and HAA-LSTM training are tested. Finally, the results of the two models are fused by arithmetic averaging the results to obtain the final results. Our experiments were performed on 1 NVIDIA RTX 3090 GPU. The AdamW optimizer was used with a batch size of 4 and a learning rate of 1e-5.

### 3.2. Ablation Study

This section describes ablation experiments to assess the effect of different training methods on the results from a specific perspective.

**Analysis of the number of training iterations.** Table 1 shows the effect of different number of iterations during training. The experiments show that the SPL, SR and GP metrics of the models are improved as the number of training rounds increases. When the number of iterations is greater than 200,000, the model shows overfitting.

| Methods | Unseen Testing | | |
|---|---|---|---|
| | SPL | SR | GP |
| HAA-Transformer(140000itre) | 12.61 | 14.05 | 54.59 |
| HAA-Transformer(20000iter) | 12.64 | 15.03 | 55.39 |
| HAA-Transformer(260000iter) | 11.74 | 14.27 | 53.51 |

Table 1 shows the effect of different number of iterations

**Effect of multiple model fusion.** In Table 2, model fusion of two models, HAA-Transformer and HAA-LSTM, yields better results compared to single model training. There is a significant improvement in the GP metrics, which helps the UAV to move towards its destination.

| Methods | Unseen Testing | | |
|---|---|---|---|
| | SPL | SR | GP |
| HAA-Transformer | 12.64 | 15.03 | 55.39 |
| HAA-LSTM | 12.67 | 14.37 | 53.66 |
| Fusion | 12.87 | 15.16 | 60.39 |

Table 2 shows the effect of fusing multiple models

### 3.3. Challenge Results

Table 3 shows our final results for the 2023 VDN challenge. Our method outperforms the original HAA-LSTM model in both SR and SPL evaluation metrics, increasing the GP from 54.6 to 60.4.

| Methods | Unseen Testing | | |
|---|---|---|---|
| | SPL | SR | GP |
| HAA-Transformer | 12.61 | 15.72 | 53.68 |
| HAA-LSTM | 12.91 | 14.05 | 54.59 |
| Ours | 12.87 | 15.16 | 60.39 |

Table 3 shows our final results in AVDN challeng.

### 4. Conclusions

This paper proposes a deep learning-based approach to aerial conversation navigation, which utilizes Transformer's powerful sequence modeling capabilities, global information perception capabilities and attention mechanism to solve this multi-modal task. We propose an effective method for fusing training Human Attention-Assisted Transformer (HAA-Transformer) model and Human Attention-Assisted LSTM (HAA-LSTM) model to achieve the prediction of navigation routing points and human attention. The method not only achieves higher SR and SPL metrics, but also a 7% improvement in GP metrics compared to the baseline model.

### References

[1] Fan, Y., Chen, W., Jiang, T., Zhou, C., Zhang, Y., and Wang, X. E., "Aerial Vision-and-Dialog Navigation", <i>arXive-prints</i>,2022. doi:10.48550/arXiv.2205.12219.

[2] Alexander Ku, Peter Anderson, Roma Patel, Eugene Ie, and Jason Baldridge. 2020. Room-across-room: Multilingual vision-and-language navigation with dense spatiotemporal grounding. arXiv preprint arXiv:2010.07954.

[3] Angel Chang, Angela Dai, Thomas Funkhouser, Maciej Halber, Matthias Niessner, Manolis Savva, Shuran Song, Andy Zeng, and Yinda Zhang. 2017. Matterport3d: Learning from rgb-d data in indoor environments. arXiv preprint arXiv:1709.06158.

[4] Jacob Krantz, Erik Wijmans, Arjun Majumdar, Dhruv Batra, and Stefan Lee. 2020. Beyond the nav-graph:

Vision-and-language navigation in continuous environments. In European Conference on Computer Vision, pages 104–120. Springer.

[5] Vihan Jain, Gabriel Magalhaes, Alexander Ku, Ashish Vaswani, Eugene Ie, and Jason Baldridge. 2019.

[6] An Yan, Xin Eric Wang, Jiangtao Feng, Lei Li, and William Yang Wang. 2020. Cross-lingual visionlanguage navigation.

[7] Keji He, Yan Huang, Qi Wu, Jianhua Yang, Dong An, Shuanglin Sima, and Liang Wang. 2021. Landmarkrxr: Solving vision-and-language navigation with fifine-grained alignment supervision. In NeurIPS.

[8] Dipendra Misra, Andrew Bennett, Valts Blukis, Eyvind Niklasson, Max Shatkhin, and Yoav Artzi. 2018. Mapping instructions to actions in 3d environments with visual goal prediction. In Proceedings of the 2018 Conference on Empirical Methods in Natural Language Processing, pages 2667–2678.

[9] Howard Chen, Alane Suhr, Dipendra Misra, Noah Snavely, and Yoav Artzi. 2019. Touchdown: Natural language navigation and spatial reasoning in visual street environments. In 2019 IEEE/CVF Conference on Computer Vision and Pattern Recognition (CVPR), pages 12530–12539.

[10] Piotr Mirowski, Andras Banki-Horvath, Keith Anderson, Denis Teplyashin, Karl Moritz Hermann, Mateusz Malinowski, Matthew Koichi Grimes, Karen Simonyan, Koray Kavukcuoglu, Andrew Zisserman, et al. 2019. The streetlearn environment and dataset. arXiv preprint arXiv:1903.01292.

[11] Karl Moritz Hermann, Mateusz Malinowski, Piotr Mirowski, Andras Banki-Horvath, Keith Anderson, and Raia Hadsell. 2020. Learning to follow directions in street view. In AAAI Conference on Artififi-cial Intelligence.

[12] Arun Balajee Vasudevan, Dengxin Dai, and LucVan Gool. 2021. Talk2nav: Long-range vision-andlanguage navigation with dual attention and spatial memory. International Journal of Computer Vision, 129(1):246–266.

[13] Yue Fan, Shilei Chu, Wei Zhang, Ran Song, and Yibin Li. 2020. Learn by observation: Imitation learning for drone patrolling from videos of a human navigator. In IEEE/RSJ International Conference on Intelligent Robots and Systems (IROS), pages 5209–5216.

[14] Lyujie Chen, Feng Liu, Yan Zhao, Wufan Wang, Xiaming Yuan, and Jihong Zhu. 2020. Valid: A comprehensive virtual aerial image dataset. In 2020 IEEE International Conference on Robotics and Automation (ICRA), pages 2009–2016. IEEE.

[15] Peter Anderson, Qi Wu, Damien Teney, Jake Bruce, Mark Johnson, Niko Sünderhauf, Ian Reid, Stephen ould, and Anton Van Den Hengel. 2018. Visionand-language navigation: Interpreting visuallygrounded navigation instructions in real environments. In Proceedings of the IEEE conference on omputer vision and pattern recognition, pages 3674–3683.

[16] Thomas Wolf, Lysandre Debut, Victor Sanh, Julien Chaumond, Clement Delangue, Anthony Moi, Pierric Cistac, Tim Rault, Rémi Louf, Morgan Funtowicz, Joe Davison, Sam Shleifer, Patrick von Platen, Clara Ma, Yacine Jernite, Julien Plu, Canwen Xu, Teven Le Scao, Sylvain Gugger, Mariama Drame, Quentin Lhoest, and Alexander M. Rush. 2020. Transformers: State-of-the-art natural language processing. In Proceedings of the 2020 Conference on Empirical Methods in Natural Language Processing: System Demonstrations, pages 38–45, Online. Association for Computational Linguistics.